\pdfoutput=1

\documentclass[11pt]{article}

\usepackage[final]{ACL2023}

\usepackage{color}
\usepackage{times}
\usepackage{latexsym}
\usepackage{graphicx}
\usepackage[T1]{fontenc}

\usepackage[utf8]{inputenc}

\usepackage{microtype}

\usepackage{inconsolata}
\usepackage{amsmath}

\usepackage{multirow} 
\usepackage{array}
\usepackage{xcolor, soul}
\usepackage{arydshln}
\usepackage{makecell}

\title{Large Language Models Can Learn Temporal Reasoning}

\author{Siheng Xiong$^1$\thanks{\ \ Equal contribution.}, Ali Payani$^2$\footnotemark[1], Ramana Kompella$^2$, Faramarz Fekri$^1$ \\
  $^1$Georgia Institute of Technology \quad $^2$Cisco Research\\
  \texttt{sxiong45@gatech.edu} \quad 
  \texttt{\{apayani, rkompell\}@cisco.com} \quad \\ \texttt{faramarz.fekri@ece.gatech.edu} \\
}

\begin{document}
\maketitle
\begin{abstract}
While large language models (LLMs) have demonstrated remarkable reasoning capabilities, they are not without their flaws and inaccuracies. Recent studies have introduced various methods to mitigate these limitations. Temporal reasoning (TR), in particular, presents a significant challenge for LLMs due to its reliance on diverse temporal concepts and intricate temporal logic.
In this paper, we propose TG-LLM, a novel framework towards language-based TR. 
Instead of reasoning over the original context, we adopt a latent representation, temporal graph (TG) that enhances the learning of TR.
A synthetic dataset (TGQA), which is fully controllable and requires minimal supervision, is constructed for fine-tuning LLMs on this text-to-TG translation task. 
We confirmed in experiments that the capability of TG translation learned on our dataset can be transferred to other TR tasks and benchmarks. On top of that, we teach LLM to perform deliberate reasoning over the TGs via Chain-of-Thought (CoT) bootstrapping and graph data augmentation. We observed that those strategies, which maintain a balance between usefulness and diversity, bring more reliable CoTs and final results than the vanilla CoT distillation.\footnote{Code and data are available at \url{https://github.com/xiongsiheng/TG-LLM}.}
\end{abstract}

\section{Introduction}

As one of the fundamental abilities, temporal reasoning (TR) plays an important role in human perception. It is not just about understanding basic concepts such as ordering or duration; it extends to more intricate aspects, e.g., task planning or causal relation discovery. Recently, large language models (LLMs) ~\citep{ouyang2022training, openai2023gpt4, touvron2023llama} have emerged with some reasoning capabilities ~\citep{huang2023reasoning}. However, there is observation that they still can not perform TR sufficiently well ~\citep{wang2023tram, chu2023timebench, qiu2023large}, preventing their applications from solving complex real-world problems. 
In particular, TR requires a combination of various skills including mathematical and logical reasoning as well as commonsense knowledge ~\citep{zhou2019going, vashishtha-etal-2020-temporal, qin2021timedial}.

\begin{figure}[t]
  \centering
  \includegraphics[width=0.49\textwidth]{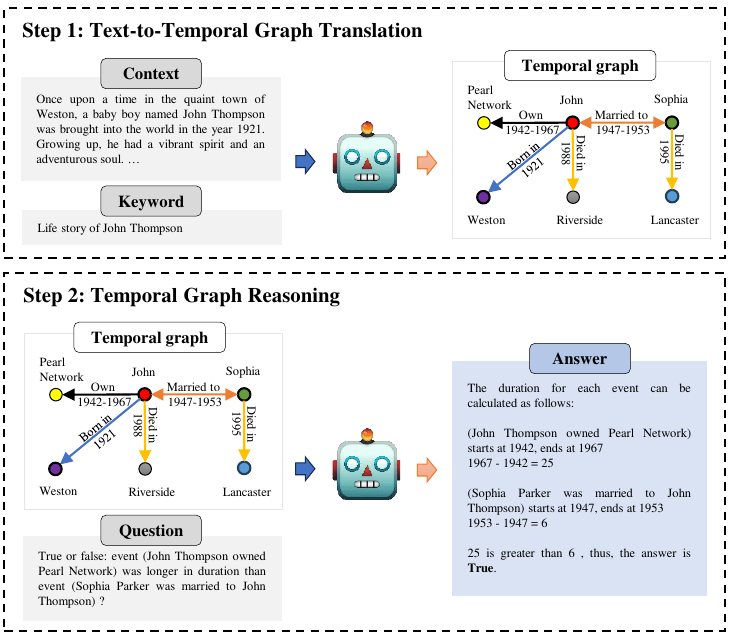}
  \caption{Our framework (TG-LLM) performs temporal reasoning in two steps: 1) Text-to-Temporal Graph translation: generate (relevant) temporal graph given the context and keyword (extracted from questions); 2) Temporal Graph Reasoning: perform Chain-of-Thought reasoning over the temporal graph.}
\end{figure}


Recent works mainly adopt general approaches to investigate and improve the TR capability of LLMs. For example, ~\citep{wang2023tram, chu2023timebench, qiu2023large} benchmark the leading LLMs on different TR tasks with standard Input/Output prompt, few-shot in-context learning (ICL) and Chain-of-Thought (CoT) reasoning ~\citep{wei2022chain}. Similarly, ~\citep{wei2023menatqa} designs several specific types of prompts as prompt tuning. ~\citep{li2023unlocking, yuan2023future} introduce pre-defined Python programs/rule-based templates to perform supervised fine-tuning (SFT). In addition, ~\citep{tan2023benchmarking, tan2023robust} adopt some extra strategies, which include specific pre-training, instruction tuning and reinforcement learning. 

Despite the effectiveness of such methods, they either ignore or not explicitly involve the intrinsic nature of TR.
Humans perform complex TR on a timeline of events which are aligned with the entities and relations. These temporal concepts (e.g., ordering, duration, frequency, typical time) are then rigorously defined based on the timeline information. 
In other words, the aligned timeline (more generally, the temporal graph, TG) serves as a latent representation to help humans understand the patterns in TR. 
However, due to the lack of ground truth, the high-quality TG translation is a challenging task for most TR benchmarks.
To solve this problem, we propose a synthetic dataset (TGQA), which is fully controllable and requires minimal supervision. We demonstrate the capability of TG translation learned on our dataset can be transferred to other TR tasks and benchmarks.

Given a reliable TG, the key challenges of teaching TR to LLMs include: (1) How can one introduce the necessary arithmetic and commonsense knowledge involved in TR? Prior work ~\citep{lewis2021retrievalaugmented} shows that explicitly introducing knowledge into context enhances the performance of LLMs. In this paper, we first identify all the valid time expressions, and then generate related knowledge (e.g., temporal relation and time gap between the timestamps, and the relative order of the gaps). (2) How can one teach LLM to perform deliberate reasoning? Generally, there exist two roadmaps: (i) translating natural language into logical statements, and using external symbolic engine for reasoning ~\cite{pan2023logiclm}; (ii) using LLMs directly as the reasoning engine ~\citep{zhu2023large}. For (i), the difficulty lies in accurate translation ~\cite{yang2023harnessing} and the limited expressive power of formal logic. For (ii), there is no guarantee for the correctness of generated intermediate steps especially with insufficient training data ~\citep{yang2023neurosymbolic}. In this paper, we adopt (ii) with the proposed bootstrapping method to generate reliable intermediate steps for supervised fine-tuning. We further improve the model performance with graph data augmentation, which mitigates the data deficiency in TR tasks.

To be specific, our contributions are summarized as follows:
\begin{itemize}
    \item We propose a new paradigm, TG-LLM, for language-based TR. In this framework, we first translate the context into a latent representation (temporal graph), and then perform reasoning on it. Extensive experiments prove that 
    our novel approach results in superior performance compared to the baselines.
    \item We design two approaches including Chain-of-Thought bootstrapping and graph data augmentation to teach LLM to generate consistent and faithful CoTs, which brings better performance than the vanilla CoT distillation.
    \item We present a pipeline to create a synthetic dataset (TGQA) for question answering that requires TR. It is fully controllable and requires minimal supervision for text-temporal graph alignment. We show in experiments that fine-tuning on our dataset benefits LLM on other TR tasks and benchmarks.
\end{itemize}

\begin{table}[t]
\small
\centering
\renewcommand{\arraystretch}{1.2}
\begin{tabular}{l}
\hline
\textbf{Temporal Graph [sub; rel; obj; start/end; time]:}\\
\hdashline
{[1]} \textcolor{brown}{(John Thompson was born in Weston) }{starts at }\textcolor{brown}{1921;}\\
{[2]} \textcolor{brown}{(John Thompson owned Pearl Network)} {starts at }\textcolor{brown}{1942;}\\
{[3]} \textcolor{brown}{(Sophia Parker was married to John Thompson) }{starts}\\
\textcolor{black}{at} \textcolor{brown}{1947;} {[4]} \textcolor{brown}{ (John Thompson was married to Sophia Parker) }\\
\textcolor{black}{starts at }\textcolor{brown}{1947;} {[5]} \textcolor{brown}{ (Sophia Parker was married to John}\\
\textcolor{brown}{Thompson) }{ends at }\textcolor{brown}{ 1953;} {[6]} \textcolor{brown}{ (John Thompson was married}\\
\textcolor{brown}{to Sophia Parker) }{ends at }\textcolor{brown}{1953;} {[7]} \textcolor{brown}{ (John Thompson owned }\\
\textcolor{brown}{Pearl Network) }{ends at }\textcolor{brown}{ 1967;} {[8]} \textcolor{brown}{ (John Thompson died in}\\
\textcolor{brown}{Riverside) }{starts at }\textcolor{brown}{ 1988;} {[9]} \textcolor{brown}{ (Sophia Parker died in}\\
\textcolor{brown}{Lancaster) }{starts at }\textcolor{brown}{ 1995.}\\
\hline
\textbf{Graph-based Story (from GPT-3.5):} \\
\hdashline
\textcolor{cyan} {Once upon a time in the quaint town of Weston, a baby}\\
\textcolor{cyan} {boy named John Thompson was brought into the world} \\
\textcolor{cyan} {in the year 1921. Growing up, he had a vibrant spirit} \\
\textcolor{cyan} {and an adventurous soul. $\cdots$} \\
\hline
\textbf{Graph-based QAs (from rule-based Python script):}\\
\hdashline
\textbf{Q1:} \textcolor{black}{Which event started first, }\textcolor{blue}{ (John Thompson owned}\\
\textcolor{blue}{Pearl Network)} \textcolor{black}{or} \textcolor{blue}{(John Thompson was married to} \\
\textcolor{blue}{Sophia Parker)}{?}\\
\textbf{A1:} \textcolor{violet}{(John Thompson owned Pearl Network).}\\
\\
\textbf{Q2:} \textcolor{black}{True or false: event }\textcolor{blue}{ (John Thompson owned Pearl}\\
\textcolor{blue}{Network) } {was longer in duration than event } \textcolor{blue}{(Sophia} \\
\textcolor{blue}{Parker was married to John Thompson)}{?}\\
\textbf{A2:} \textcolor{violet}{True.}\\
\textcolor{blue}{$\cdots$}
\\\hline
\end{tabular}
\caption{Each sample from TGQA dataset is in the form of (temporal graph, story, questions, answers).}
\label{table:TGQA_sample}
\end{table}

\begin{table*}[t!]
\small
\centering
\renewcommand{\arraystretch}{1.2}
\begin{tabular}{l|l|l}
\hline
\textbf{Reasoning Type} & \textbf{Question} & \textbf{Answer}\\
\hline
\cline{2-3} Sequencing & Given the following <N> events: <Event\_A>, <Event\_B>, <Event\_C>, & <Event\_A>/<Event\_B>/\\ 
&<Event\_D>, $\cdots$, which event is the first/second/third/fourth/$\cdots$ one in & <Event\_C>/<Event\_D>/\\
& the chronological order? & $\cdots$\\
\hline
Duration & How long did the event <Event\_A> last?
&<Duration\_of\_Event\_A> \\
\hline
Temporal& How much time passed between the start of <Event\_A> and the start of &<Gap\_between\_Event\_A\\
Relation &<Event\_B>? & \_and\_Event\_B\_startTime>\\
\cline{2-3}&What happened right before/after <Event\_A> started? &<Event\_B> right before/\\
&&after <Event\_A>\\
\hline
Facts &When did the <Event\_A> occur? &<Event\_A\_startTime> \\
Extraction & &\\
\hline
Simultaneity & True or false: <Event\_A> and <Event\_B> happened at the same year?
&True / False \\
\cline{2-3}&True or false: <Event\_A> was still happening when <Event\_B> started? &True / False\\
\hline
Comparative & True or false: <Event\_A> was longer in duration than <Event\_B>?&True / False\\
Analysis & &\\
\hline
\end{tabular}
\caption{Reasoning types with the corresponding questions and answers in TGQA.}
\label{table:TGQA_question_types}
\end{table*}

\section{Dataset Construction}

In this section, we present the construction pipeline for TGQA dataset that is fully controllable and requires minimal supervision for text-temporal graph alignment. Compared with existing datasets ~\citep{chen2021dataset, tan2023benchmarking}, we have the ground-truth timelines and more diverse categories and types of TR questions (Table \ref{table:TGQA_question_types}). More importantly, the pipeline can be used for various scenarios and tasks. We first split the large temporal knowledge graph, YAGO11k \citep{dasgupta2018hyte}, into subgraphs with a restriction on the number of events, and anonymize the entities to avoid data leakage. Then each story is translated from the subgraph by GPT-3.5 ~\citep{ouyang2022training}. By using some rule-based templates, we obtain reliable question and answer (QA) pairs from the graph. Finally, to reduce the noise introduced from the misalignment between the subgraph and generated story, we propose a semi-automatic verification method. 


\noindent \textbf{Step 1: Graph Splitting \& Anonymization.\ } Existing temporal knowledge graphs \citep{leetaru2013gdelt, dasgupta2018hyte, garcia-duran-etal-2018-learning} usually have a large size. To facilitate the learning process, we split YAGO11k into subgraphs for story generation. 
Specifically, given a certain entity, we find its neighbors within three hops, and extract all the events happening between them. 
Since we hope LLMs to do reasoning instead of memorization, it is ensured that no overlapping exists between the events in training, validation and test sets. 
Additionally, we notice the data leakage problem of LLMs, i.e., prior knowledge of the test data has been implicitly obtained from the extensive pre-training. Thus, an anonymization strategy, i.e., changing entity names into random ones of the same type, is adopted. For each relation, we generate a global mapping of entity names with GPT-3.5. 
To avoid confusion, we adopt obscure names that do not exist in YAGO11k.

\noindent \textbf{Step 2: Graph-based Open QA Creation.\ } In TGQA, each sample is in the form of (temporal graph, story, questions, answers) (Table \ref{table:TGQA_sample}). Based on the given subgraph, we generate a story and multiple QAs. We first ask GPT-3.5 to write a story based on the subgraph with the requirement to include all the events. It is observed that GPT-3.5 tends to ignore the end time of some events in the created story. To solve this problem, we separate the start and end time of the same event in the prompt. On the other hand, to obtain a comprehensive benchmark, we consider all types of temporal reasoning, which include sequencing, duration, frequency, simultaneity, temporal relation, comparative analysis and facts extraction. Given these categories, we design multiple question types (Table \ref{table:TGQA_question_types}) with a rule-based Python script to generate the corresponding Qs and As.

\noindent \textbf{Step 3: Quality Control.\ } In TGQA, noise might be introduced from the misalignment between the given subgraph and LLM-generated story. To address this problem, we propose a semi-automatic verification method. Fully manual inspection is expensive, but by first utilizing LLM we can narrow down potential errors, and only manually inspect the set of unanswered questions by LLM. Specifically, given a generated story, GPT-3.5 is queried on the time of each event in the graph. If it cannot give the correct answer, we consider the event as possibly missing in the story, which requires further manual verification. 
We proved the effectiveness of our semi-automatic pipeline with fully manual verification of the test stories. Note that supervision is only required for story-TG alignment verification, since all the QAs are generated from rules. We show dataset statistics in Appendix \ref{appendix:dataset_statistics}, and all the prompts involved in Appendix \ref{appendix:example_prompt}.


\section{TG-LLM}

Motivated by human perception, we propose a new paradigm called TG-LLM. We first translate the text into a temporal graph (TG), and then guide the LLM to perform deliberate reasoning on it.

\subsection{Text-to-TG Translation}

Although LLM (with ICL) might have such capability, we observed a misalignment between the generated TG and pre-defined QAs, i.e, LLM making mistakes or focusing on irrelevant events. Since TG serves as the foundation of the following deliberate reasoning process, we provide a pipeline for high-quality TG data generation to fine-tune the LLM.


\noindent \textbf{Ground-truth TG Generation.} For some datasets such as TGQA, a verified TG (corresponding to the story) is provided. We can directly fine-tune the LLM on this task with the ground truth. 
However, for most real-world applications, the biggest challenge is the lack of ground truth TG. We provide a pipeline for high-quality TG data generation. We first extract all the entities and relations from the QAs to help LLM focus on specific events. We then identify all the valid time expressions in the story. We provide these information to LLM for better alignment in TG construction, and verify the generated TG in a semi-automatic way.

{(1) Entity \& Relation Extraction:} To obtain the entity and relation in pre-defined QAs, we consider two strategies: parsing with rules or using GPT-3.5. Rule-based parsing, if applicable, is more efficient and reliable, which is prioritized in our experiments. 

{(2) Temporal Info Identification:} 
To identify all the valid time expressions, we first use GPT-3.5 to extract from the story, and then adopt a rule-based Python script for invalid output filtering and normalization. 
Besides bringing better alignment, the pre-processing of time expressions facilitates the introduction of external knowledge (their temporal relations and time gaps) into TR. We proved in experiments that explicitly introducing this information further improves model performance. 


{(3) TG Construction \& Verification:} We generate the TG using GPT-3.5 with ICL. Specifically, we provide several in-context demonstrations which include a story and extracted entities, relations and time expressions as input and the corresponding TG as output (Table \ref{tab:graph_construction}). 
For those stories without explicit time expressions, we skip Step (2), and ask GPT-3.5 to provide the temporal order of events. Similar to TGQA, we verify the generated TG by first asking GPT-3.5 the pre-defined QAs, and then manually checking the alignment if GPT-3.5 fails.

\begin{figure}[!t]
    \centering
    \includegraphics[width=0.42\textwidth]{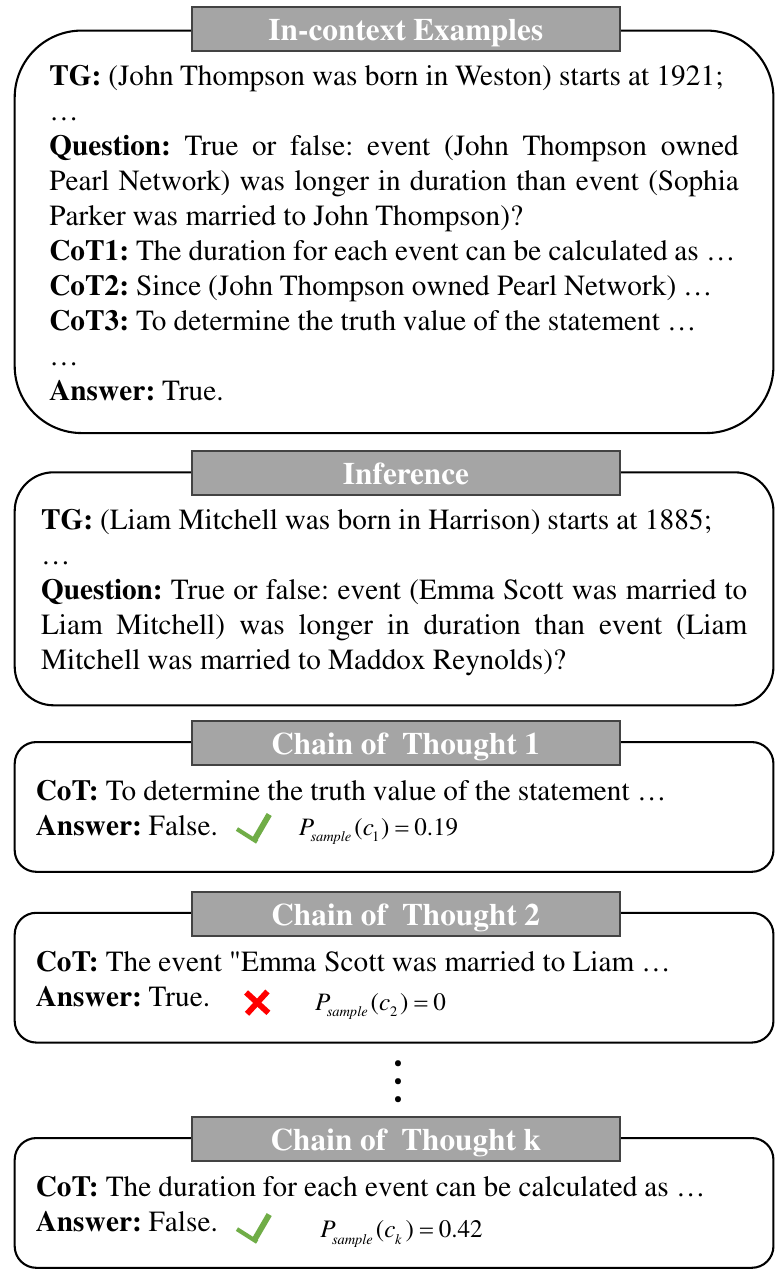}
  \caption{In Chain-of-Thought (CoT) bootstrapping, we only accept CoTs that lead to correct final answers and sample them according to their contrastive learning scores to balance usefulness and diversity.}
\end{figure}

\noindent \textbf{Supervised Fine-tuning.} 
We first conducted experiments to decide the best in-context format of TG. It is found out that providing a chronological list of events helps LLM perform better TR. In addition, it offers advantages to separate the start and end time of the same event. With these steps, the TG in context is transferred into a timeline with the alignment between entities, relations and times. We perform supervised fine-tuning (SFT) using Llama-2 model ~\citep{touvron2023llama2} with Low-Rank Adaptation (LoRA) ~\citep{hu2021lora}. The input and output of the model are the story and aligned timeline, respectively. In experiments, we observe benefits from the SFT on our dataset to other temporal reasoning tasks.

\subsection{Temporal Graph Reasoning}

Given the generated TGs, we teach LLM deliberate reasoning with SFT enhanced by CoT bootstrapping and graph data augmentation.

\subsubsection{Bootstrapping Chain of Thoughts}

It is observed that SFT on reliable CoTs brings better reasoning performance than that on standard Input/Output prompts ~\citep{wang2023scott, ho2023large}. Since asking humans to create the CoT data is not scalable, we use LLM to replace humans. 
This task is non-trivial for reasoning over knowledge graphs ~\citep{saparov2023language}.
In this section, we propose a bootstrapping pipeline, i.e., given a query, using LLM to generate several CoTs and selecting them as training data with a weighted sampling strategy. Compared with the conventional Best-of-N sampling, our proposal allows more training data diversity.  

Specifically, we first prepare ICL examples for high-quality CoT generation. To facilitate the learning, the pre-defined QAs are classified into different categories $\{Q_{1}, Q_2, \cdots$, $Q_M\}$. For each category $Q_i$, we randomly choose $N_i$ samples $\{(g_j, e_j, q_j, a_j)\}^{N_i}_{j=1}$, where $g$, $e$, $q$, $a$ denote TG, external knowledge, question and answer, respectively, and ask both GPT-3.5 and GPT-4 to provide diverse CoTs $\{c_j\}^{N_i}_{j=1}$. These CoTs will then be manually verified as ICL examples. Given a new training sample $(g_{j^{\prime}}, e_{j^{\prime}}, q_{j^{\prime}})$, we bootstrap $K$ CoTs with final answers $\{(c_{j^{\prime},k}, \hat{a}_{j^{\prime},k})\}^{K}_{k=1}$ from GPT-3.5.
We first refuse the CoTs leading to incorrect answers, i.e., $\hat{a}_{j^{\prime},k}\neq a^*_{j^{\prime}}$, where $a^*_{j^{\prime}}$ denotes the correct answer. 
For the accepted CoTs, we consider a weighted sampling strategy to balance usefulness and diversity. The sampling probability $P_{sample}(\cdot)$ is based on a contrastive learning $score$ (Eq. \ref{eq:sample_prob}).
The $score$ design, inspired by ~\citep{wang2023scott}, considers both normalized probability $P(\cdot)$ of the correct answer and plausibility growth $G(\cdot)$ (Eq. \ref{eq:contrastive_learning_score}). The definition of $G(\cdot)$ and $P(\cdot)$ are given in Eq. \ref{eq:plausibility growth} and Eq. \ref{eq:perplexity}, respectively.
{
\fontsize{10pt}{12pt}
\begin{align}
& P_{sample}(c_k) = \operatorname{softmax}(score(c_k)) \label{eq:sample_prob}\\
& score(c_k) = \log P\left(a^* \mid q^{\dagger}, c_k\right) + \gamma G\left(c_k\right) \label{eq:contrastive_learning_score}\\
& G\left(c_k\right)=\log \frac{P\left(a^* \mid q^{\dagger}, c_k\right)}{\Bar{P}_{\{a^{\prime} \in A^{\prime}\}}\left(a^{\prime} \mid q^{\dagger}, c_k\right)}\label{eq:plausibility growth}\\
& \log P\left(a \mid q^{\dagger}, c_k\right) = \frac{1}{|a|} \sum_{l=0}^{|a|} \log P\left(t_l \mid q^{\dagger}, t_{<l}\right) \label{eq:perplexity}
\end{align}
}



\noindent where $c_k$ denotes the current CoT in consideration; $q^{\dagger} := \{g,e,q\}$; $a^{\prime}$ denotes a certain wrong answer from the candidate set $A^{\prime}$; weight $\gamma$ is a hyperparameter; $t_l$ denotes the $l$-th token in $a$, and $t_{<l}$ denotes the sequence of tokens before $t_l$.

\begin{figure}[!b]
    \centering
    \includegraphics[width=0.48\textwidth]{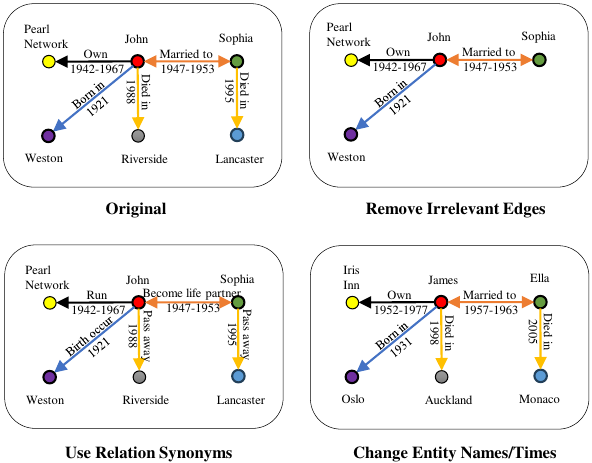}
  \caption{We further boost the model performance with several graph data augmentation strategies: remove irrelevant edges, use relation synonyms and change entities/times.}
  \label{figure:data_aug}
\end{figure}


\subsubsection{Graph Data Augmentation}


Compared with other tasks, reasoning suffers more from data insufficiency since more information (such as evidence, arguments, and logics) are involved in the intermediate steps ~\citep{huang2023reasoning}. To address this issue, we propose several graph data augmentation strategies (Figure \ref{figure:data_aug}).

Our framework involves two steps: text-to-TG translation and temporal graph reasoning. 
Notably, for the reasoning part, the model is trained over ground-truth/verified TGs but infers on the estimated graphs. This discrepancy actually hurts the robustness of reasoning. Thus, we introduce some disturbances on the TG during training. 
Note that the type of disturbances should be carefully designed in order to avoid confusing the LLM. 
We first investigate two types of disturbances: (i) remove irrelevant edges (Eq. \ref{eq:irr_fact}) and (ii) replace edges by using relation synonyms (Eq. \ref{eq:rel_mapping}).
An edge (event) is considered irrelevant if not involved in both QA and CoT.
{
\fontsize{11pt}{12pt}
\begin{align}
& F_{irr} \in \{ F : P(c|g_{\setminus\{F\}},e,q) \approx P_0 \}_{F \in g} \label{eq:irr_fact} \\   
& F^{\prime} \in \{ f_R(F) : P(c|f_R(g),e,q) \approx P_0 \}_{F \in g} \label{eq:rel_mapping}
\end{align}
}

\noindent where $P_0 := P(c|g,e,q)$ denotes the original conditional probability, irrelevant event $F_{irr}$ will be randomly removed from $g$, and disturbed event $F^{\prime}$ is obtained from the global mapping of relation names $f_R(\cdot)$.

Furthermore, to make sure LLM learns the underlying logic of TR instead of just memorizing semantic information,
we introduce another two types of disturbances: (iii) globally map all the entity names in training data to some random names of the same type (Eq.\ref{eq:ent_and_time_mapping})
, and (iv) change the times based on a global offset (Eq.\ref{eq:ent_and_time_mapping}). 
For each relation, we generate these random entity names with GPT-3.5 by providing several examples of existing names. 

{
\fontsize{11pt}{12pt}
\begin{equation}
\begin{aligned}
& g^{\prime} = f_T(f_E(g)), \ e^{\prime} = f_T(e), \\   
& q^{\prime} = f_T(f_E(q)), \ c^{\prime} = f_T(f_E(c)),\\
& a^{\prime} = f_T(f_E(a))
\end{aligned}
\label{eq:ent_and_time_mapping}
\end{equation}
}

\noindent where $f_E(\cdot), f_T(\cdot)$ denote the global mapping of entity names and time changing, respectively.

\section{Experiments}

We aim to answer the following research questions in our experiments: (1) Can our strategies (CoT bootstrapping and graph data augmentation) bring more reliable reasoning over TGs? (2) Can our two-step framework lead to better TR performance? (3) Do these learned capabilities of TR generalize to other tasks?

\subsection{Experimental Setup} We demonstrate TG-LLM is a general framework by applying it to, besides TGQA, the two existing datasets, TimeQA ~\citep{chen2021dataset} and TempReason ~\citep{tan2023benchmarking}, which are constructed using Wikipedia articles, excerpts, and summaries. Examples from other datasets are listed in Appendix \ref{appendix:experiment_details}.
We thoroughly evaluate our framework on all the datasets with a combination of the metrics, i.e., token-level F1, exact match (EM) and perplexity-based accuracy (Acc). 
Besides choosing F1 and EM, which are two basic metrics for span-based QA tasks, we consider Acc for LLM evaluation, i.e., selecting from a candidate set the final answer with the lowest perplexity as the prediction. The rationale and detailed construction of the candidate set for all datasets are listed in Appendix \ref{appendix:experiment_details}.

We primarily compare our framework
with the leading LLMs, i.e., Llama2 \cite{touvron2023llama2}, GPT-3.5 \citep{ouyang2022training} and GPT-4 \citep{openai2023gpt4}. Specifically, we ran the Llama2 family on local machines (4-bit quantization for Llama2-70B due to limited computational resources), and used APIs provided by OpenAI for GPT models. We evaluated their few-shot ICL performance on the test set. To prove the effectiveness of our method, we also show the model performance with SFT on standard Input/Output prompts (SP) and CoTs from GPT-3.5.
The model versions and prompt templates are provided in Appendix \ref{appendix:experiment_details} and \ref{appendix:example_prompt}, respectively. We also include the T5-based models ~\citep{tan2023benchmarking, yang2023textittime} for a comprehensive comparison. Results on TimeQA and TempReason reported in the original papers are used, and these models are fine-tuned and evaluated on TGQA. 

\begin{figure}[!b]
    \centering
    \includegraphics[width=0.45\textwidth]{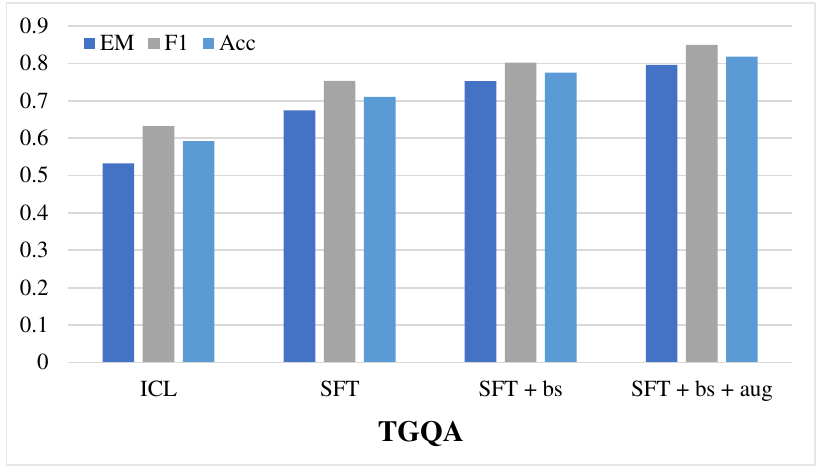}
  \caption{Performance comparison between different CoT generation strategies on TGQA.}
   \label{figure:CoT_gen}
\end{figure}

\begin{table}[!b]
\small
\centering
\begin{tabular}{lcccc}
\hline
{Strategy} & {ER-T1} & {ER-T2} & {ER-T3} & {ER-T4} \\
\hline
ICL & 0.13 & 0.13 & 0.31 & 0.10 \\
SFT & 0.04 & 0.17 & 0.13 & 0.10 \\
SFT + bs & 0.06 & 0.13 & 0.03 & 0.08 \\ 
SFT + bs + aug & 0.03 & 0.05 & 0.04 & 0.05 \\\hline
\end{tabular}
\caption{Human evaluation on the generated CoTs by different strategies. ER: error rate; T1: using wrong info; T2: logical inconsistency; T3: external knowledge error; T4: temporal graph error. (Error type explanations are listed in Appendix \ref{appendix:experiment_details}.)}
\label{table:human_eval_CoTs}
\end{table}

\begin{table*}[ht]
\scriptsize
\begin{center}
\renewcommand{\arraystretch}{1.3}
\setlength{\tabcolsep}{4.7pt}
\begin{tabular}{l|ccc|ccc|ccc|ccc|ccc}
\Xhline{0.9pt}
\multicolumn{1}{l|}{\multirow{3}{*}{\textbf{Model}}}  & \multicolumn{3}{c|}{\multirow{2}{*}{\textbf{TGQA}}} &\multicolumn{6}{c|}{\textbf{TimeQA}}&\multicolumn{6}{c}{\textbf{TempReason}}\\
\multicolumn{1}{l|}{}  &\multicolumn{3}{c|}{} &\multicolumn{3}{c}{\textbf{Easy-mode}} &\multicolumn{3}{c|}{\textbf{Hard-mode}}&\multicolumn{3}{c}{\textbf{OBQA-L2}} &\multicolumn{3}{c}{\textbf{OBQA-L3}}\\
\cline{2-16}
\multicolumn{1}{l|}{} & \multicolumn{1}{c}{EM}  &\multicolumn{1}{c}{F1} &\multicolumn{1}{c|}{Acc} & \multicolumn{1}{c}{EM}  &\multicolumn{1}{c}{F1} &\multicolumn{1}{c|}{Acc} & \multicolumn{1}{c}{EM}  &\multicolumn{1}{c}{F1} &\multicolumn{1}{c|}{Acc}& \multicolumn{1}{c}{EM}  &\multicolumn{1}{c}{F1} &\multicolumn{1}{c|}{Acc}& \multicolumn{1}{c}{EM}  &\multicolumn{1}{c}{F1} &\multicolumn{1}{c}{Acc}
\\ 
\hline
T5-base$^{\dag}$  &  0.410	 &  0.608 & - & 0.600 &	0.682& - & 0.556	& 0.641& - & 0.260 & 0.450 & - & 0.238& 0.418&-\\
T5-large$^{\dag}$  &  0.548	 &  0.713 & - & 0.631 &	0.716& - & 0.595	& 0.681& - & 0.327&0.509&-&0.288&0.468&-\\
Temp-T5$^{\dag}$ &  0.640	 &  0.778 & - & - &	-& - & -	& - & -&0.318&0.496&-&0.261&0.430&-\\
REMEMO-base$^{\dag}$  &  0.435	 &  0.633 & - & 0.614 &	0.704& - & 0.582	& 0.673& -&0.336&0.516&-&0.285&0.449&-\\
REMEMO-large$^{\dag}$  &  0.461	 &  0.660 & - & 0.637 &	0.723& - & 0.605	& \textbf{0.693}& -&0.374&\textbf{0.549}&-&0.334&\textbf{0.493}&-\\
\hline
GPT-3.5 (ICL-SP)  &  0.598	 &  0.699 & - & 0.640 &	0.668& - & 0.512	& 0.506& - & 0.303 & 0.409 & - & 0.365 & 0.453 & -\\
GPT-3.5 (ICL-CoT)  & 0.706 & 0.788 & -& 0.565 & 0.603 & -& 0.436 & 0.464& - &0.340&0.478&-&0.243&0.348&-\\
GPT-4$^*$ (ICL-SP)  &  0.772 & 0.829 & -& \textbf{0.716} & \textbf{0.742} & -& 0.571 & 0.546& - & \textbf{0.454} & 0.525 & - & \textbf{0.431} & 0.485&-\\
GPT-4$^*$ (ICL-CoT)  &  \textbf{0.821} & \textbf{0.865} & - & 0.662 & 0.693& - & 0.618 & 0.636& - &0.388&0.480&-&0.352&0.447&-\\
Llama2-7B (ICL-SP)  &  0.415 & 0.596& 0.447& 0.352 & 0.408 &0.367& 0.341 & 0.404&0.354 & 0.205 & 0.344 & 0.226& 0.044& 0.084&0.153\\
Llama2-7B (ICL-CoT)  &  0.548 & 0.686& 0.578& 0.367 & 0.425& 0.393& 0.302 & 0.354&0.360 & 0.233 & 0.410 & 0.263 & 0.179 & 0.357&0.187\\
Llama2-13B (ICL-SP)  &  0.440 & 0.609 & 0.526& 0.439 & 0.493 &0.450& 0.427 & 0.481&0.437& 0.284&0.452 & 0.289& 0.189 &0.370 & 0.183\\
Llama2-13B (ICL-CoT)  &  0.628 & 0.762&0.668 & 0.518 & 0.572& 0.535& 0.434 & 0.490&0.480&0.330&0.498&0.368 &0.242&0.391&0.272\\
Llama2-70B (ICL-SP)  &  0.618 & 0.736 & 0.665 & 0.583 & 0.627& 0.631& 0.493 & 0.537&0.551&0.358&0.491&0.387&0.128&0.181&0.148\\
Llama2-70B (ICL-CoT)  &  0.761 & 0.838 & 0.796 & 0.552 & 0.612 & 0.623 & 0.447 & 0.501& 0.512&0.325&0.477&0.345&0.303&0.393&0.318\\
\hline
Llama2-13B (SFT-SP)  &   0.550 & 0.720 &0.652& 0.412 & 0.449 &0.455& 0.362 & 0.404&0.412&0.337&0.506& 0.338&0.244&0.408&0.253\\
Llama2-13B (SFT-CoT)  &  0.646 & 0.722 & 0.705& 0.550 & 0.586 & 0.564& 0.332& 0.391 & 0.379&0.256&0.433&0.281&0.285&0.409&0.305\\
Llama2-13B (SFT-TGR)  &  {0.797} & {0.850}& \textbf{0.819} & 0.664 & 0.691 &\textbf{0.673}& \textbf{0.631} & {0.664}&\textbf{0.649} & 0.424&0.522&\textbf{0.432}&0.356&0.469&\textbf{0.399}\\
\Xhline{0.9pt}
\end{tabular}
\caption{Main results using different models and strategies. We report exact match (EM), token-level F1 scores, and perplexity-based accuracy (Acc). Note: (1) Results with $^{\dag}$ are reported in the original papers. We only fine-tune and evaluate the models on our dataset. (2) Results with * are evaluated on 1000 random test samples.}
\label{table-main-results}
\end{center}
\end{table*}

\begin{figure*}[!t]
    \centering
    \includegraphics[width=\textwidth]{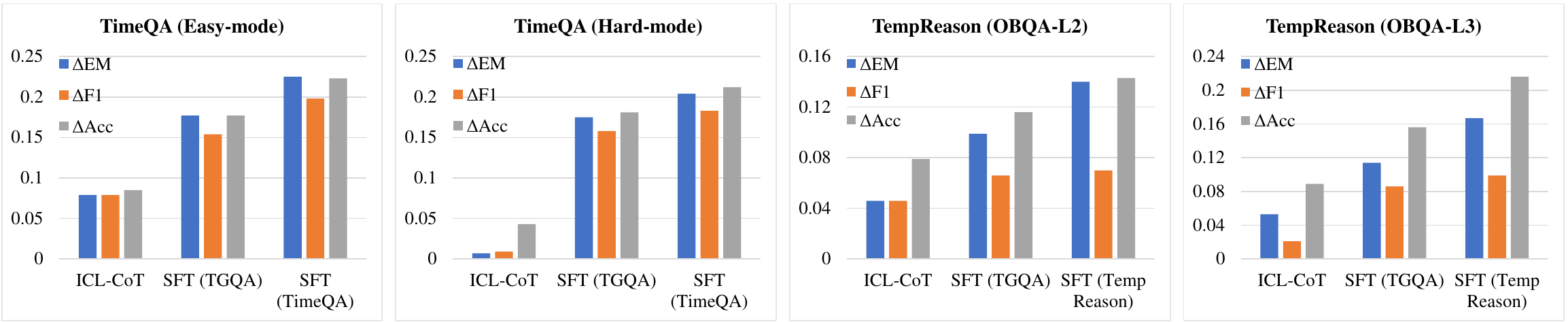}
  \caption{Performance comparison between different strategies on TimeQA and TempReason. To obtain a fair comparison, we use Llama2-13B as the base model for all strategies. The basic strategy used to calculate the performance changes is in-context learning with standard Input/Output prompt (ICL-SP).}
  \label{fig:transfer_learning}
\end{figure*}

\subsection{Implementation Details}

We use Llama2-13B as the baseline due to limited computational resources. 
We inject two adapters with selectors into the base model for the text-to-TG translation and temporal graph reasoning. The adapters are trained in parallel. For inference, we first translate the original story into a temporal graph, and then perform reasoning on it, i.e., the adapters are used in sequence. For data generation, we use GPT-3.5 for story, TG and CoT generation, and the verification of stories and TGs. We use GPT-4 to create the ICL demonstrations of CoT generation, due to its high generation quality.
All the prompt templates are given in Appendix \ref{appendix:example_prompt}.

\subsection{Main Results}

\textbf{Can our strategies bring more reliable reasoning over TGs?} We show the comparison between ICL with CoTs, SFT with CoTs, bootstrapping CoTs, and graph data augmentation on TGQA (Figure \ref{figure:CoT_gen}). 
It can be seen that LLM learns TR better with SFT than ICL. By providing CoTs with bootstrapping and graph data augmentation strategies, the model performance gets further enhanced. 
Furthermore, inspired by ~\citep{wang2023scott}, we manually check 100 CoTs generated by different strategies (Table \ref{table:human_eval_CoTs}). Evaluators are asked to classify the errors into four types (using wrong info, logical inconsistency, external knowledge error, and temporal graph error). 
It can be seen that our strategies reduce all types of error rate.


\noindent \textbf{Can our two-step framework lead to better temporal reasoning performance?} We show the comparison between different models and strategies on all datasets (Table \ref{table-main-results}). First, we observed that among all the LLMs with ICL, GPT-4 has the strongest performance as expected. For the Llama2 family, larger models have better performance due to advanced context understanding and improved generalization. We also found that CoTs not always bring better TR due to unreliable intermediate results and hallucinations. 
From the results of some alternative strategies, where SFT-SP and SFT-CoT denote supervised fine-tuning on standard Input/Output prompts and vanilla CoT distillation, respectively, we prove the effectiveness of our framework (SFT-TGR).
More importantly, our model, which is based on Llama2-13B, shows a comparable or even better performance than GPT-4 on all datasets.
It can be seen that the two-step framework brings a substantial performance improvement on different datasets. 
We hypothesize this performance improvement is because our two-step reasoning process provides an easier path toward answering the temporal questions for LLM.

\noindent \textbf{Do these learned capabilities of temporal reasoning generalize to other tasks?} We show the comparison between different strategies (ICL with SP/CoT, SFT with TGQA/original data) on the two existing datasets, TimeQA and TempReason (Figure \ref{fig:transfer_learning}). 
Our framework learns the capabilities of text-to-TG translation and temporal graph reasoning, which brings better TR. More importantly, we observed that SFT on TGQA improves the model performance compared with ICL. 
It can be concluded that these necessary capabilities in TR are generalizable to different data distributions. Since TGQA is fully controllable and requires minimal supervision, we actually provide a general and effective way of TR capability improvement.  

\subsection{Ablation Study}

We ablate different modules to see their contributions to the performance. We show the performance comparison between different configurations (Table \ref{tab:ablation_study}).
To obtain a fair comparison, we use Llama2-13B as the base model for all configurations. 
From the ablation study, we obtain some insights: (1) LLM can benefit from explicitly presented (temporal) graph which is intuitive, concise and structured. (2) Given a reliable graph, CoT bootstrapping with contrastive learning brings better performance than vanilla CoT distillation. (3) Data augmentation is necessary for LLMs to perform complex tasks such as temporal reasoning. (4) The introduction of external knowledge such as mathematics and commonsense can further augment the generation.
 
\section{Related Work}

\noindent \textbf{Language-based Temporal Reasoning.} Recently, language-based TR has gained substantial research focus ~\citep{liu2023grounding, liu2024delta, chen2024timelinebased, chen2024selfimprovement, wang2024klink, jiayang2024eventground, xia2024enhancing}. The vision here is to help LMs understand temporal concepts and logic such that they can perform more complicate tasks. 
Existing methods mainly solve this problem with time-aware language modeling. For example, ~\citep{rosin2022time, pereira2022attention, tan2023benchmarking} propose specific pre-training/fine-tuning strategies for robust TR. On the other hand, ~\citep{ning2019joint, zhou2020temporal, zhou2021temporal, yang2020improving, yang2023textittime} design auxiliary objectives to introduce external temporal knowledge.
Although these methods made some progress, representation learning for the underlying structure and logic of TR are either ignored or not explicitly involved. 

\begin{table}[!t]
\small
\centering
\begin{tabular}{lccc}
\hline
{Config.} & {EM} & {F1} & {Acc}  \\
\hline
Baseline (SFT-CoT) & 0.646 & 0.722 & 0.705 \\
TG & 0.675 & 0.754 & 0.711 \\
TG + CoT(bs) & 0.723 & 0.797 & 0.736 \\
TG + CoT(bs + aug) & 0.782 & 0.838 & 0.795  \\
TG + EK + CoT(bs) & 0.753 & 0.803 &  0.776 \\
TG + EK + CoT(bs + aug) & \textbf{0.797} & \textbf{0.850} & \textbf{0.819} \\
\hline
\end{tabular}
\caption{Ablation study results on TGQA with Llama2-13B. TG: temporal graph, CoT: chain of thought, EK: external knwoledge, bs: bootstrapping, aug: graph data augmentation.}
\label{tab:ablation_study}
\end{table} 

\noindent \textbf{Reasoning towards LMs.} Although LLM has exhibited some emergent behaviors \citep{zhang2024unlocking, lai2024language, lai2024adaptive, lyu2024taskagnostic}, it is still unknown whether they can actually perform reasoning and how strong their capability of reasoning is.
Existing methods that try to elicit or enhance reasoning can be divided into two directions: reasoning-involved modeling or hybrid methods. For example, ~\citep{wei2022chain, kojima2022large, yao2023tree} use in-context learning where some demonstrations including intermediate reasoning steps are provided in prompt. ~\citep{nye2021show, wang2023scott, ho2023large} fine-tune LMs with the intermediate thinking process before producing final answers. On the other hand, ~\citep{schick2023toolformer, kynoch2023recallm, xu2023gentopia} combine LMs with domain-specific external tools, empowering the model to perform more complex tasks that require reasoning and interactions with environment. 

\noindent \textbf{Reasoning over Knowledge Graphs.}
Knowledge graphs (KGs) as the foundation representation of semantic and symbolic reasoning have been widely adopted in the past \citep{huang2023federated, wan2024federated, li2024enhancing}. Related work includes symbolic reasoning over temporal KGs ~\citep{yang2022temporal, xiong2023tilp, xiong2024teilp}, and language-based reasoning over static KGs ~\citep{cheng2024necessary, zhang2024dietodin, liu2024evaluating, zhao2024all, xu2024generateongraph, wei2024versatile}. In this paper, we build the connection between language-based and symbolic-based TR. This connection brings the potential for extending these KG-based methods to language-based tasks. 
Different from existing methods~\citep{luo2023chatrule, zhang2023making, yuan2023future, gao-etal-2024-two}, which limit their application to certain tasks, 
our framework offers enhanced generalization and usability, largely attributed to its innovative use of text-to-graph translation as a precursor to graph reasoning.

\section{Conclusion}

TG-LLM, a novel framework for LMs, has been proposed to improve their performance on temporal reasoning. To produce reliable final answers, our framework equips LLMs with the temporal graph and intermediate reasoning steps. Extensive experiments indicate that TG-LLM achieves better performance than existing pipelines. An interesting direction for future work is to extend it to more complex applications such as inductive and abductive reasoning. Due to the graph structure and capability of deliberate reasoning, it is promising to improve the model performance on these tasks as well.

\section*{Limitations}
Graph-augmented approaches ~\citep{jin2024graph, fan2024graph, shang2024survey, he2024unigraph} including TG-LLM help language models better learn related concepts from the perspective of graph. Although we demonstrate TG-LLM has good performance on understanding temporal relations, it still needs adaptations for temporal commonsense reasoning ~\citep{zhou2019going, qin2021timedial}. Explicit in-context integration of commonsense presents opportunities for this task. Further, we mainly improve the capability of LLMs by introducing a new paradigm and providing more plausible and informative training data. There can be opportunities such as simulating an environment to provide feedback to LLMs ~\citep{hao2023reasoning}. For example, we can verify the generated TG based on prior knowledge such as the time gap between someone's birth date and death date, i.e., a person's lifespan, should fall into a certain range. In this way, we can further improve the performance.

\section*{Ethics Statement}

In this paper, we adopt YAGO11k for fine-tuning the language models. The dataset is publicly available, and is for research purposes only. We also use GPT model to generate text based on YAGO11k, for which OpenAI has been committed to addressing ethical considerations. In addition, we adopt TimeQA and TempReason for evaluation. Both datasets are publicly available, and are for research purposes only. However, they may still contain improper or harmful content. None of such content reflects the opinions of the authors.

\section*{Acknowledgements}
This work was supported by a sponsored research award by Cisco Research.

\bibliography{anthology,custom}
\bibliographystyle{acl_natbib}

\newpage

\appendix

\begin{figure}[!t]
    \centering
    \includegraphics[width=0.45\textwidth]{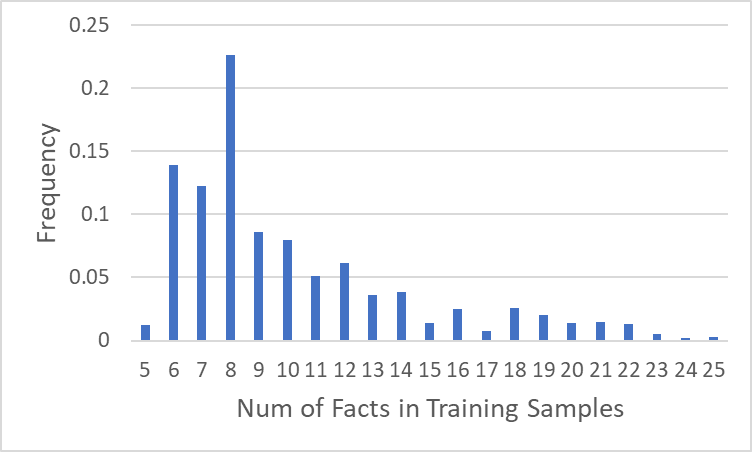}
    \includegraphics[width=0.45\textwidth]{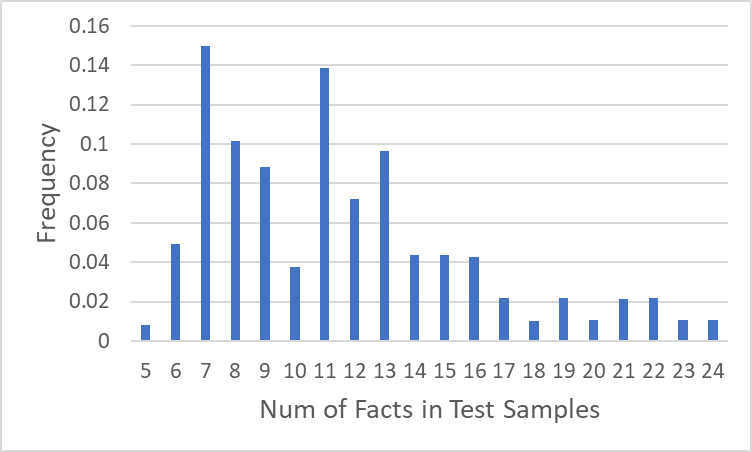}
  \caption{Number of facts distribution in TGQA.}
  \label{fig:num_facts_dist}
\end{figure}

\begin{figure}[!t]
    \centering
    \includegraphics[width=0.45\textwidth]{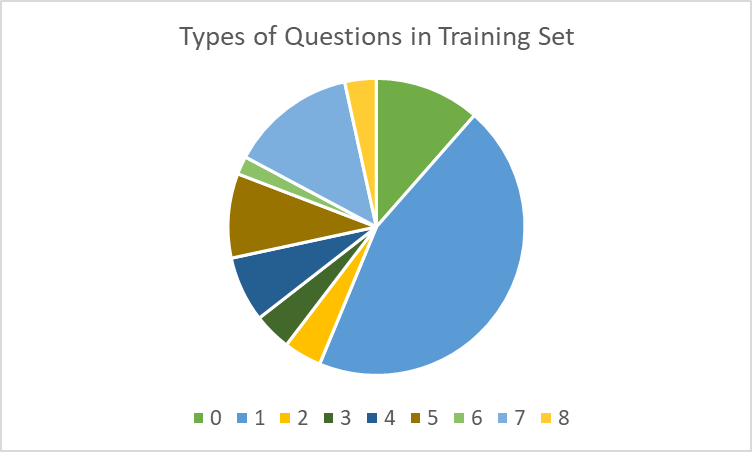}
    \includegraphics[width=0.45\textwidth]{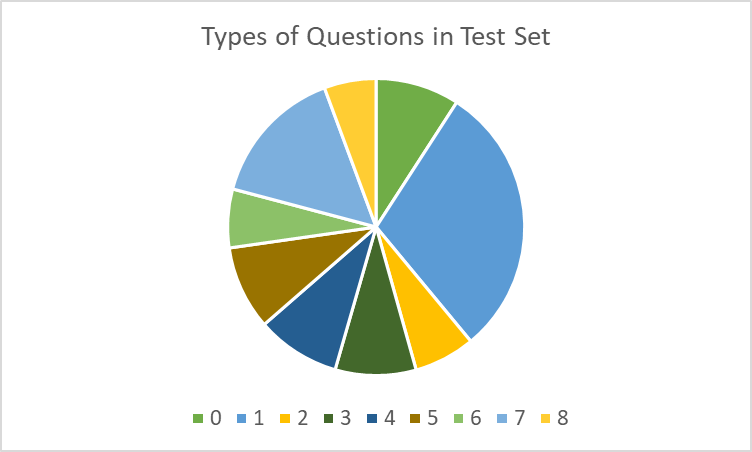}
  \caption{Types of questions distribution in TGQA.}
  \label{fig:question_type_dist}
\end{figure}

\section{Dataset Statistics of TGQA}
\label{appendix:dataset_statistics}
In this section, we provide statistics for our TGQA dataset. We obtain 400 samples for training, 100 for validation and another 100 for test, with about 30 QA pairs in a single sample. We show the distribution of the number of facts (in one sample) in training (with validation) and test set (Figure \ref{fig:num_facts_dist}). It determines the complexity of the TR tasks to some extent. We also show the distribution of question types in training (with validation) and test set (Figure \ref{fig:question_type_dist}), where Q0: "Which event occurred first, <Event\_A> or <Event\_B>?", Q1: "Given the following <N> events: <Event\_A>, <Event\_B>, <Event\_C>, <Event\_D>, $\cdots$, which event is the first/second/third/fourth/$\cdots$ one in the chronological order?", Q2: "How long did the event <Event\_A> last?", Q3: "", Q4: "How much time passed between the start of <Event\_A> and the start of <Event\_B>?", Q5: "What happened right before/after <Event\_A> started?", Q6: "When did the <Event\_A> occur?", Q7: "True or false: <Event\_A> and <Event\_B> happened at the same year?", Q8: "True or false: <Event\_A> was still happening when <Event\_B> started?". Note that we give eight question categories in Table \ref{table:TGQA_question_types}, since in a strict sense Q0 and Q1 can be considered as the same type. Among all the question types, Q1 has the largest portion since it has multiple variants. Similarly, Q7 with two variants also has a larger portion. To mitigate question category imbalance, we first calculate the metrics for each category, and then use the average as the final scores. Additionally, we show examples for the global mapping for entity names (Table \ref{table:global_ent_mapping}) and external knowledge (Table \ref{table:external_knowledge}) used in our dataset.

\begin{table}[!t]
\small
\centering
\renewcommand{\arraystretch}{1.2}
\begin{tabular}{cc}
\hline
\multicolumn{2}{c}{\textbf{Entity Name Mapping}}\\
\textbf{Ori} & \textbf{New}\\
\hline
Beverly\_Adams & Isabella\_Thompson\\
Edmonton & Lancaster\\
Al\_Gore & Chris\_Evans\\
Carrara\_Stadium & Maplewood\_Arena\\
Bend\_Sinister\_(novel) & Lakefield\_Chronicles\_(novel)\\
$\cdots$ & $\cdots$\\
\hline
\end{tabular}
\caption{We use a global mapping for entity names from GPT-3.5 to avoid data leakage.}
\label{table:global_ent_mapping}
\end{table} 


\begin{table}[!t]
\small
\centering
\renewcommand{\arraystretch}{1.2}
\begin{tabular}{l}
\hline
\multicolumn{1}{c}{\textbf{External Knowledge}}\\
\hline
1885 before 1893 before 1916 before 1918 before 1922 \\
before 1928 before 1941\\
1918 - 1916 = 2\\
1928 - 1893 = 35\\
1928 - 1922 = 6\\
1941 - 1918 = 23\\
2 < 6 < 23 < 35\\
\hline
\end{tabular}
\caption{We integrate the necessary mathematics and commonsense in context as external knowledge for TR.}
\label{table:external_knowledge}
\end{table}

\section{Experiment Details}
\label{appendix:experiment_details}
In this section, we present more experiment details for further research. We include graph data augmentation examples, model versions and evaluation tasks \& metrics.

\begin{table}[!t]
\small
\centering
\renewcommand{\arraystretch}{1.2}
\begin{tabular}{l}
\hline
\textbf{Temporal Graph:}\\
\hdashline
(\textbf{Use relation synonyms/Remove irrelevant edges})\\
{[1]} \textcolor{brown}{(John Thompson was born in Weston) }{starts at }\textcolor{brown}{1921;}\\
{[2]} \textcolor{brown}{(John Thompson} \hl{run} \textcolor{brown}{Pearl Network)} {starts at }\textcolor{brown}{1942;}\\
{[3]} \textcolor{brown}{(Sophia Parker} \hl{and} \textcolor{brown}{John Thompson} \hl{became life}\\
\hl{partner}\textcolor{brown}{)} \textcolor{black}{starts at} \textcolor{brown}{1947;} {[4]} \textcolor{brown}{(John Thompson} \hl{and} \textcolor{brown}{Sophia}\\
\textcolor{brown}{Parker} \hl{became life partner}\textcolor{brown}{)} \textcolor{black}{starts at }\textcolor{brown}{1947;} {[5]} \textcolor{brown}{(Sophia}\\
\textcolor{brown}{Parker} \hl{and} \textcolor{brown}{John Thompson} \hl{became life partner}\textcolor{brown}{)}\\
{ends at }\textcolor{brown}{ 1953;} {[6]} \textcolor{brown}{(John Thompson} \hl{and} \textcolor{brown}{Sophia Parker)}\\
\hl{became life partner}\textcolor{brown}{)} {ends at }\textcolor{brown}{1953;} {[7]} \textcolor{brown}{(John Thompson}\\
 \hl{run} \textcolor{brown}{Pearl Network) }{ends at }\textcolor{brown}{ 1967.}\\
\hline
\textbf{Graph-based QAs:}\\
\hdashline
(\textbf{No need for change})\\
\textbf{Q:} \textcolor{black}{True or false: event }\textcolor{blue}{ (John Thompson owned Pearl}\\
\textcolor{blue}{Network) } {was longer in duration than event } \textcolor{blue}{(Sophia} \\
\textcolor{blue}{Parker was married to John Thompson)}{?}\\
\textbf{CoT:} \textcolor{gray}{The duration for each event can be} \\
\textcolor{gray}{calculated as follows:}\\
\textcolor{gray}{(John Thompson owned Pearl Network) starts at}\\ \textcolor{gray}{1942, ends at 1967, 1967 - 1942 = 25}\\
\textcolor{gray}{(Sophia Parker was married to John Thompson) starts} \\\textcolor{gray}{at 1947, ends at 1953, 1953 - 1947 = 6}\\
\textcolor{gray}{25 is greater than 6 , thus, the answer is True.}\\
\textbf{A:} \textcolor{violet}{True.}\\
\hline
\textbf{Temporal Graph:}\\
\hdashline
(\textbf{Change entities/times})\\
{[1]} \textcolor{brown}{(}\hl{James Brown} \textcolor{brown}{was born in }\hl{Oslo}{) }{starts at} \hl{1931}\textcolor{brown}{;}\\
{[2]} \textcolor{brown}{(}\hl{James Brown} \textcolor{brown}{owned} \hl{Iris Inn}\textcolor{brown}{)} {starts at }\hl{1952}\textcolor{brown}{;}\\
{[3]} \textcolor{brown}{(}\hl{Ella Perry} \textcolor{brown}{was married to }\hl{James Brown}\textcolor{brown}{) }{starts}\\
\textcolor{black}{at} \hl{1957}\textcolor{brown}{;} {[4]} \textcolor{brown}{ (}\hl{James Brown} \textcolor{brown}{was married to }\hl{Ella Perry}\textcolor{brown}{) }\\
\textcolor{black}{starts at }\hl{1957}\textcolor{brown}{;} {[5]} \textcolor{brown}{ (}\hl{Ella Perry} \textcolor{brown}{ was married to} \hl{James}\\
\hl{Brown}\textcolor{brown}{) }{ends at }\hl{1963}\textcolor{brown}{;} {[6]} \textcolor{brown}{ (}\hl{James Brown} \textcolor{brown}{ was married}\\
\textcolor{brown}{to }\hl{Ella Perry}\textcolor{brown}{) }{ends at }\hl{1963}\textcolor{brown}{;} {[7]} \textcolor{brown}{ (}\hl{James Brown} \textcolor{brown}{ owned }\\
\hl{Iris Inn}{) }{ends at }\hl{1977}\textcolor{brown}{;} {[8]} \textcolor{brown}{ (}\hl{James Brown} \textcolor{brown}{ died in}\\
\hl{Auckland}\textcolor{brown}{) }{starts at }\hl{1998}\textcolor{brown}{;} {[9]} \textcolor{brown}{ (}\hl{Ella Perry} \textcolor{brown}{ died in}\\
\hl{Monaco}\textcolor{brown}{) }{starts at }\hl{2005}\textcolor{brown}{.}\\
\hline
\textbf{Graph-based QAs:}\\
\hdashline
(\textbf{Change entities/times})\\
\textbf{Q:} \textcolor{black}{True or false: event }\textcolor{blue}{ (}\hl{James Brown} \textcolor{blue}{owned} \hl{Iris Inn}\textcolor{blue}{)}\\
{was longer in duration than event } \textcolor{blue}{(}\hl{Ella Perry} \textcolor{blue}{was}\\
\textcolor{blue}{married to }\hl{James Brown}\textcolor{blue}{)}{?}\\
\textbf{CoT:} \textcolor{gray}{The duration for each event can be} \\
\textcolor{gray}{calculated as follows:}\\
\textcolor{gray}{(}\hl{James Brown} \textcolor{gray}{owned} \hl{Iris Inn}\textcolor{gray}{) starts at }\hl{1952}\textcolor{gray}{, ends at}\\
\hl{1977}\textcolor{gray}{, }\hl{1977} \textcolor{gray}{-} \hl{1952} \textcolor{gray}{= 25}\\
\textcolor{gray}{(}\hl{Ella Perry} \textcolor{gray}{was married to }\hl{James Brown}\textcolor{gray}{) starts at} \\
\hl{1957} \textcolor{gray}{, ends at }\hl{1963}\textcolor{gray}{, }\hl{1963} \textcolor{gray}{-} \hl{1957} \textcolor{gray}{ = 6}\\
\textcolor{gray}{25 is greater than 6 , thus, the answer is True.}\\
\textbf{A:} \textcolor{violet}{True.}\\
\hline
\end{tabular}
\caption{We propose several graph data augmentation strategies for reasoning over temporal knowledge graph. The original TG is shown in Table \ref{table:TGQA_sample}. We highlight the changed information.}
\label{table:example_data_aug}
\end{table} 

\noindent \textbf{Graph Data Augmentation.\ }
We show with an example our graph data augmentation strategies in Table \ref{table:example_data_aug}. For TGs with relation synonym replacement or irrelevant edges removal, there is no need for change on graph-based QAs. To contrast, for TGs with entity names/times mapping, we need to change with the corresponding entity names/times in graph-based QAs to ensure the consistency.

\noindent \textbf{Evaluation Tasks \& Metrics.\ } 
Besides TGQA, we consider in experiments the two existing datasets TimeQA ~\citep{chen2021dataset} and TempReason ~\citep{tan2023benchmarking}. Specifically, 
TimeQA contains two difficulty levels (Table \ref{table:eamples_TimeQA}). The easy-level split tends to be the information extraction task while the hard-level split involves understanding the relation between different temporal expressions. On the other hand, TempReason contains three levels (L1: Time-Time Relation, L2: Time-Event Relation, L3: Event-Event Relation) and three settings (OBQA: open-book QA, CBQA: close-book QA, ReasonQA: facts-based QA) (Table \ref{table:eamples_TempReason}). We observed that some stories in TempReason are incomplete which partially leads to the low accuracy of LLMs.

We evaluate our framework on all the datasets with the metrics of token-level F1, exact match (EM) and perplexity-based accuracy (Acc). 
F1 and EM are two basic metrics for span-based QA tasks. However, the free-form prediction of LLMs might hurt their performance under these generation-based metrics. To solve this problem, we introduce perplexity-based accuracy, i.e., selecting from a candidate set the final answer with the lowest perplexity.
For questions with multiple correct answers, we follow the strategy proposed in ~\citep{chen2021dataset} to get the best result among them. Since there are multiple question categories in TGQA, we first calculate the metrics for each category, and then use the average as final scores to mitigate question category imbalance.



\begin{table}[!t]
\small
\centering
\renewcommand{\arraystretch}{1.2}
\begin{tabular}{l}
\hline
\multicolumn{1}{c}{\textbf{TimeQA}}\\
\hline
George Washington (February 22, 1732 – December 14, \\
1799) was an American Founding Father, military offic-\\
er, politician and statesman who served as the first $\cdots$\\
\\
Questions (Easy-mode):\\
What position did George Washington hold in June 17-\\
75?\\
What position was held by George Washington between \\
1778 and 1788?\\
$\cdots$\\
\\
Questions (Hard-mode):\\
George Washington took which position before 1778?\\
What was George Washington’s position in early 1780s?\\
$\cdots$\\
\\
\hline
\end{tabular}
\caption{Example questions of two difficulty levels in TimeQA. Easy-mode: the query time expression is explicitly mentioned in the story. Hard-mode: obtaining the answer needs inference based on the temporal relation between the query time expression and the one mentioned in the story.}
\label{table:eamples_TimeQA}
\end{table}

\begin{table}[!t]
\small
\centering
\renewcommand{\arraystretch}{1.2}
\begin{tabular}{l}
\hline
\multicolumn{1}{c}{\textbf{TempReason}}\\
\hline
Lionel Andrés "Leo" Messi (born 24 June 1987) is an Ar-\\
gentine professional footballer who plays as a forward for \\
and captains both Major League Soccer club $\cdots$\\
\\
Questions (L1):\\
What is the year after 2010?\\
$\cdots$\\
\\
Questions (L2):\\
What team did Leo Messi play for in 2010?\\
$\cdots$\\
\\
Questions (L3):\\
What team did Leo Messi play for after Barcelona?\\
$\cdots$\\
\\
\hline
\end{tabular}
\caption{Example questions of three difficulty levels under the OBQA setting in TempReason. L1: Time-Time Relation, L2: Time-Event Relation, L3: Event-Event Relation.}
\label{table:eamples_TempReason}
\end{table}

\begin{table*}[!t]
\scriptsize
\begin{center}
\renewcommand{\arraystretch}{1.3}
\setlength{\tabcolsep}{4.7pt}
\begin{tabular}{l|cccccccccc}
\Xhline{0.9pt}
\multicolumn{1}{l|}{\multirow{2}{*}{\textbf{Model}}}  & \multicolumn{10}{c}{\multirow{1}{*}{\textbf{TGQA}}}\\
\cline{2-11}
\multicolumn{1}{l|}{} & \multicolumn{1}{c}{Q0}  &\multicolumn{1}{c}{Q1} &\multicolumn{1}{c}{Q2} & \multicolumn{1}{c}{Q3}  &\multicolumn{1}{c}{Q4} &\multicolumn{1}{c}{Q5} & \multicolumn{1}{c}{Q6}  &\multicolumn{1}{c}{Q7} &\multicolumn{1}{c}{Q8} &\multicolumn{1}{c}{Total}
\\
\hline
GPT-3.5$^\ddag$ (0-shot SP)  & 0.801 & 0.435 & 0.660 & 0.355 & 0.930 & 0.955 & 0.710 & 0.315 & 0.617 & 0.642\\
GPT-3.5 (1-shot SP)  & 0.842 & 0.411 & 0.656 & 0.555 & 0.924 & 0.650 & 0.500 & 0.353 & 0.495 & 0.598\\
GPT-3.5$^\ddag$ (0-shot CoT)  & 0.930 &0.565 &0.680 &0.665& 0.920 &0.945& 0.695 &0.395 &0.596 &0.710\\
GPT-3.5 (1-shot CoT)  & 0.884&  0.457&  0.783 & 0.791&  0.921 & 0.832 & 0.755 & 0.357&  0.574&  0.706\\
GPT-4$^{\ddag*}$ (0-shot SP)  & 0.801 & 0.689 & 0.920 & 0.620 & \textbf{0.930} & 0.940 & 0.758 & 0.575  & 0.737  & 0.776 \\
GPT-4$^{*}$ (1-shot SP)  & 0.820 & 0.660 & 0.870 &  0.630 &  0.910 & \textbf{0.990} & 0.760 &  0.610 & 0.700 & 0.772\\
GPT-4$^{\ddag*}$ (0-shot CoT)  & 0.950 & 0.692 & 0.901 & \textbf{0.881} & 0.919 & 0.980 & 0.778 & 0.647 & \textbf{0.858} & \textbf{0.848}\\
GPT-4$^{*}$ (1-shot CoT)  & 0.930 & \textbf{0.758} & \textbf{0.903} &  0.869 & 0.892 & 0.920 & 0.758 & \textbf{0.667} & 0.680 & 0.821\\
Llama2-7B (1-shot SP) & 0.554 & 0.216 & 0.163 & 0.147 & 0.353 & 0.924 & 0.683 & 0.115 & 0.579 & 0.415\\
Llama2-7B (1-shot CoT) & 0.660 & 0.264 & 0.484 & 0.603 & 0.693 & 0.947 & 0.632 & 0.079 & 0.574 & 0.548\\
Llama2-13B (1-shot SP) &  0.442 & 0.341 & 0.353 & 0.192 & 0.686 & 0.640 & 0.561 & 0.163 & 0.580 & 0.440\\
Llama2-13B (1-shot CoT) & 0.597 & 0.263 & 0.670 & 0.678 & 0.881 & 0.947 & 0.736 & 0.258 & 0.622 & 0.628\\
Llama2-70B (1-shot SP)  & 0.752 & 0.447 & 0.520 & 0.349 & 0.941 & 0.891 & 0.665 & 0.413 & 0.585 & 0.618\\
Llama2-70B (1-shot CoT)  & \textbf{0.980} & 0.490 & 0.878 & 0.832 & 0.944 & 0.878 & \textbf{0.807} & 0.294 & 0.750 & 0.761\\
\hline
Llama2-13B (1-shot SFT - TG)  & 0.878 & 0.515 & 0.656 & 0.866 & 0.835 & 0.845 & 0.726 & 0.369 & 0.383 & 0.675 \\
Llama2-13B (1-shot SFT
- TG + CoT(bs))  & \textbf{0.947} & 0.732 & \textbf{0.747} & 0.825 & \textbf{0.931} & \textbf{0.987} & 0.495 & 0.466 & 0.649 & 0.753 \\
Llama2-13B (1-shot SFT 
- TG + CoT(bs + aug)) & 0.931 & 0.710 & 0.729 &\textbf{0.860} &0.927& 0.977& 0.627 & \textbf{0.534} & \textbf{0.739} & 0.782\\
Llama2-13B (1-shot SFT - TG + EK + CoT(bs + aug))  & 0.944 & \textbf{0.775} & 0.729 & 0.849 & 0.924 & 0.980 & \textbf{0.783} & 0.506 & 0.681 & \textbf{0.797}\\
\Xhline{0.9pt}
\end{tabular}
\caption{Fine-grained results on TGQA using different models and strategies. We report exact match (EM) as the performance metric. Note: (1) Results with * are evaluated on 1000 random test samples. (2) Results with $\ddag$ are parsed by GPT-3.5 during evaluation (might introduce errors).}
\label{table-TGQA-results}
\end{center}
\end{table*}

Human evaluation on CoT generation found out four types of error in total (Table \ref{table:human_eval_CoTs}). T1 means that LLM uses wrong information such as wrong start/end time during reasoning. T2 suggests that LLM makes logical errors, e.g., mentioning "Event A ends before Event B starts" in CoT but determining the statement "Event A was still happening when Event B starts" to be true. T3 denotes that LLM makes errors on external knowledge, e.g., claiming that "the date 1978 is after 1983". T4 indicates that there exists errors in the extracted TG that lead to the wrong conclusion.

\noindent \textbf{Candidate Answer Generation.} We involve candidate answers in the calculation of Acc.
For TempReason, the authors provide negative answers since they perform time-sensitive reinforcement learning. That is, they use the score of the correct answers and wrong answers from the language model as reward to further finetune the model parameter. For TimeQA, we generate the candidates in the following way. Given a story, there are multiple related questions which share the same subject/object entity and relation. The correct answer changes with the query time. For example, "What position did George Washington hold in 1777/1790/1799?" Answer: "Commander in Chief/Presidency/Chancellor". We collect the answers of these related questions as candidates. More generally, if there are no such related questions, we will use all the entities in the corresponding TG as candidates.

\noindent \textbf{Model Versions.\ } The versions of the LLMs used in our experiments are listed below. For the Llama2 family, all the model weights can be downloaded from the platform of Hugging Face. For the GPT models, all the model weights can be accessed through the OpenAI APIs.

\begin{itemize}
    \item Llama2-7B (meta-llama/Llama-2-7b-hf)
    \item Llama2-13B (meta-llama/Llama-2-13b-hf)
    \item Llama2-70B (meta-llama/Llama-2-70b-hf)
    \item GPT-3.5 (gpt-3.5-turbo)
    \item GPT-4 (gpt-4-1106-preview)
\end{itemize}


\section{Example Prompts}
\label{appendix:example_prompt}

In this section, we show example prompts used in our framework. Specifically, Table \ref{table:example_story_gen} shows an example of the graph-based story generation in TGQA; Table \ref{table:tory_TG_align} presents an example of the automatic story-temporal graph alignment verification in TGQA; Table \ref{table:example_temporal_info_ident} provides an example of the temporal info identification in TimeQA; Table \ref{tab:graph_construction} illustrates an example of the graph construction in TimeQA; Table \ref{example_TG_QA_align} gives an example of the automatic temporal graph-QA alignment verification in TimeQA; Table \ref{example_CoT_bs} depicts an illustrative case of the CoTs bootstrapping in TGQA.

\section{Fine-grained Results of TGQA}

We provide the fine-grained results of TGQA in Table \ref{table-TGQA-results}. In TGQA, there are nine types of questions as explained in Appendix \ref{appendix:dataset_statistics}. We show all the models with different strategies. Zero-shot performance is also considered to investigate the effect of in-context examples. Note that we do not include zero-shot performance of the Llama2 family since they are not fine-tuned on instructions, i.e., we can not obtain valid zero-shot learning results. For zero-shot learning results, the format of generation is not guaranteed. The original accuracy is very low since rule-based parser cannot handle it. Instead, we use GPT-3.5 as parser to extract the answer from the generation (might introduce errors). Also, we only adopt one example in context due to context length limitation (long-context LLM performance could increase with more in-context demonstrations). 

It can be seen that in-context demonstrations are not necessary for powerful pre-trained models such as GPT-3.5 and GPT-4. Undoubtedly, GPT-4 obtains the best in-context learning performance on most categories. However, for some categories, the best performance of GPT-4 is around 0.6 or 0.7, which is not sufficiently well. On the other hand, we show that our strategies improve the performance in a progressive manner. Note that our model achieves the best overall performance by improving on all the categories. Specifically, deliberate reasoning with CoT bootstrapping brings better performance for most categories. Furthermore, graph data augmentation and external knowledge further enhance the capability of the model while preserving the existing good performance on those categories.

\begin{table*}[ht]
\small
\centering
\renewcommand{\arraystretch}{1.2}
\begin{tabular}{l}
\hline 
(Molly Adams was born in Seattle) starts at 1896 \\
(Liam Thomas Dawson was born in Seattle) starts at 1896 \\
(Molly Adams was married to Liam Thomas Dawson) starts at 1920 \\
(Liam Thomas Dawson was married to Molly Adams) starts at 1920 \\
(Liam Thomas Dawson won prize James Parker Prize for Advanced Biomedical\\
Studies) starts at 1946 \\
(Liam Thomas Dawson won prize Champion Award in Biology or Science) starts at 1947 \\
(Liam Thomas Dawson won prize Oakley Smith Prize) starts at 1948 \\
(Liam Thomas Dawson won prize Member of the National Academy) starts at 1950 \\
(Molly Adams was married to Liam Thomas Dawson) ends at 1957 \\
(Liam Thomas Dawson was married to Molly Adams) ends at 1957 \\
(Liam Thomas Dawson died in Kingsville, Texas) starts at 1984 \\
\\
Based on the above facts, generate a short story for me.\\
\hline
\end{tabular}
\caption{Example of the graph-based story generation in TGQA.}
\label{table:example_story_gen}
\end{table*}

\begin{table*}[ht]
\small
\centering
\renewcommand{\arraystretch}{1.2}
\begin{tabular}{l}
\hline 
Once upon a time, in the vibrant city of Seattle, two remarkable individuals, Molly\\
Adams and Liam Thomas Dawson, were born in the year 1896. Little did they know that \\
their destinies were entwined from the very beginning. \\
\\
Growing up in the same neighborhood, Molly and Liam developed a deep friendship  \\
that blossomed into something more as they entered adulthood. In the year 1920, their\\ love story officially began as they exchanged vows and embarked on a journey of \\companionship that would last for decades.\\
\\
$\cdots$\\
\\
When did the event (Molly Adams was married to Liam Thomas Dawson) end?\\
\hline
\end{tabular}
\caption{Example of the story-temporal graph alignment verification in TGQA.}
\label{table:tory_TG_align}
\end{table*}

\begin{table*}[ht]
\small
\centering
\renewcommand{\arraystretch}{1.2}
\begin{tabular}{l}
\hline 
Knox Cunningham\\
\\
Sir Samuel Knox Cunningham, 1st Baronet, QC (3 April 1909 – 29 July 1976) was a \\
Northern Irish barrister, businessman and politician. As an Ulster Unionist politician at\\
a time when the Unionists were part of the Conservative Party, he was also a significant\\
figure in United Kingdom politics as Parliamentary Private Secretary to Harold\\
Macmillan. His nephew was Sir Josias Cunningham.\\
\\
$\cdots$\\
\\
Extract all the time expressions such as 'June 1994', '1973', 'late 1980s'.\\
\hline
\end{tabular}
\caption{Example of the temporal info identification in TimeQA.}
\label{table:example_temporal_info_ident}
\end{table*}

\begin{table*}[ht]
\small
\centering
\renewcommand{\arraystretch}{1.2}
\begin{tabular}{l}
\hline 
Knox Cunningham\\
\\
Sir Samuel Knox Cunningham, 1st Baronet, QC (3 April 1909 – 29 July 1976) was a \\
Northern Irish barrister, businessman and politician. As an Ulster Unionist politician at\\
a time when the Unionists were part of the Conservative Party, he was also a significant\\
figure in United Kingdom politics as Parliamentary Private Secretary to Harold\\
Macmillan. His nephew was Sir Josias Cunningham.\\
\\
$\cdots$\\
\\
Construct a timeline for Knox Cunningham's position. You should only consider these\\
time points (3 April 1909, 1930s, 1939, 1942, 1943, 1945, 1947, 1949, 1954, 29 July 1976).\\
\hline
\end{tabular}
\caption{Example of the temporal graph construction in TimeQA.}
\label{tab:graph_construction}
\end{table*}

\begin{table*}[ht]
\small
\centering
\renewcommand{\arraystretch}{1.2}
\begin{tabular}{l}
\hline 
3 April 1909: Knox Cunningham was born.\\
1930s: He studied law and began his legal career.\\
1935: On 2 July 1935, he married Dorothy Enid Riley JP.\\
1939: Cunningham was called to the Bar by the Middle Temple.\\
1942: He was called to the Bar in Northern Ireland.\\
1943: Cunningham fought the Belfast West by-election.\\
1945: He contested the same seat in the general election.\\
1947: Knox Cunningham became involved with the World Alliance of YMCAs.\\
1949: He became Chairman of the National Council of the YMCA.\\
1954: Cunningham was elected to Orpington Urban District Council.\\
29 July 1976: Knox Cunningham passed away.\\
\\
Which position did Knox Cunningham hold from May 1955 to Apr 1956?\\
\hline
\end{tabular}
\caption{Example of the temporal graph-QA alignment verification in TimeQA.}
\label{example_TG_QA_align}
\end{table*}

\begin{table*}
\small
\centering
\renewcommand{\arraystretch}{1.2}
\begin{tabular}{l}
\hline 
Example:\\
Timeline:\\
(Liam Mitchell was born in Harrison) starts at 1885\\
(Maddox Reynolds was born in Glenville) starts at 1893\\
(Liam Mitchell was married to Emma Scott) starts at 1916\\
(Emma Scott was married to Liam Mitchell) starts at 1916\\
(Liam Mitchell was married to Emma Scott) ends at 1918\\
(Emma Scott was married to Liam Mitchell) ends at 1918\\
(Maddox Reynolds was married to Liam Mitchell) starts at 1922\\
(Liam Mitchell was married to Maddox Reynolds) starts at 1922\\
(Maddox Reynolds was married to Liam Mitchell) ends at 1928\\
(Liam Mitchell was married to Maddox Reynolds) ends at 1928\\
(Liam Mitchell died in Boston) starts at 1941\\
(Emma Scott died in Oceanview) starts at 1984\\
\\
Question: True or false: event (Liam Mitchell was married to Maddox Reynolds) was \\
longer in duration than event (Emma Scott was married to Liam Mitchell)?\\
\\
Useful information:\\
1885 before 1893 before 1916 before 1918 before 1922 before 1928 before 1941\\
\hline
\end{tabular}
\end{table*}

\begin{table*}
\small
\centering
\renewcommand{\arraystretch}{1.2}
\begin{tabular}{l}
\hline
1918 - 1916 = 2\\
1928 - 1893 = 35\\
1928 - 1922 = 6\\
1941 - 1918 = 23\\
2 < 6 < 23 < 35\\
\\
Answer: Let's think step by step.\\
\\
To determine whether the statement is true or false, we need to compare the durations\\
of (Liam Mitchell was married to Maddox Reynolds) and (Emma Scott was married to\\
Liam Mitchell) . The duration for each event can be calculated as follows:\\
\\
(Liam Mitchell was married to Maddox Reynolds) :\\
(Liam Mitchell was married to Maddox Reynolds) starts at 1922\\
(Liam Mitchell was married to Maddox Reynolds) ends at 1928\\
1928 - 1922 = 6\\
\\
(Emma Scott was married to Liam Mitchell) :\\
(Emma Scott was married to Liam Mitchell) starts at 1916\\
(Emma Scott was married to Liam Mitchell) ends at 1918\\
1918 - 1916 = 2\\
\\
Now, we compare the duration of the two events: 6 is greater than 2 , thus, the answer is True .\\
\\
Test:\\
Timeline:\\
(Chris Evans was born in Bristol, Connecticut) starts at 1948\\
(Oliver Jensen was born in Bristol, Connecticut) starts at 1948\\
(Chris Evans was married to Oliver Jensen) starts at 1970\\
(Oliver Jensen was married to Chris Evans) starts at 1970\\
(Chris Evans created Pearl Network) starts at 2005\\
(Chris Evans won prize Victory Achievement Award) starts at 2007\\
(Chris Evans was married to Oliver Jensen) ends at 2010\\
(Oliver Jensen was married to Chris Evans) ends at 2010\\
(Chris Evans created Pearl Network) ends at 2013\\
\\
Question: True or false: event (Chris Evans owned Pearl Network) was longer in \\
duration than event (Oliver Jensen was married to Chris Evans)?\\
\\
Useful information:\\
1948 before 1970 before 2005 before 2007 before 2010\\
1970 - 1948 = 22\\
2007 - 1948 = 59\\
2007 - 2005 = 2\\
2010 - 1970 = 40\\
2 < 22 < 40 < 59\\
\\
Answer: Let's think step by step.\\
\hline
\end{tabular}
\caption{Example of the CoT bootstrapping in TGQA.}
\label{example_CoT_bs}
\end{table*}

\end{document}